\documentclass[runningheads,a4paper]{llncs}

\usepackage{amssymb}
\usepackage{graphicx}

\usepackage{url}
\newcommand{\keywords}[1]{\par\addvspace\baselineskip
\noindent\keywordname\enspace\ignorespaces#1}

\begin{document}

\pdfoutput = 1

\mainmatter  

\title{Video Saliency Detection by 3D Convolutional Neural Networks}

\titlerunning{Video Saliency Detection by 3D Convolutional Neural Networks}

%
%
\author{Guanqun Ding, Yuming Fang}
\authorrunning{Guanqun Ding et al.}

\institute{School of Information Technology,\\
Jiangxi University of Finance and Economics, Nanchang 330013, China\\
\url{leo.fangyuming@gmail.com}}

%
%

\toctitle{Lecture Notes in Computer Science}
\tocauthor{Authors' Instructions}
\maketitle

\begin{abstract}
Different from salient object detection methods for still images, a key challenging for video saliency detection is how to extract and combine spatial and temporal features. In this paper, we present a novel and effective approach for salient object detection for video sequences based on 3D convolutional neural networks. First, we design a 3D convolutional network (Conv3DNet) with the input as three video frame to learn the spatiotemporal features for video sequences. Then, we design a 3D deconvolutional network (Deconv3DNet) to combine the spatiotemporal features to predict the final saliency map for video sequences. Experimental results show that the proposed saliency detection model performs better in video saliency prediction compared with the state-of-the-art video saliency detection methods.
\end{abstract}

\section{Introduction}

Saliency detection, which attempt to automatically predict conspicuous and attractive regions/objects in a given image or video, has been actively studied in the field of image processing and computer vision recently. Always considered as a preprocessing procedure, saliency detection can effectively filter out redundant visual information yet preserve important regions, and it has been widely used in a variety of computer vision tasks, such as object recognition~\cite{1Rutishauser2004}, image retargeting~\cite{22Fang2012} and summarization~\cite{3Simakov2008}.

Over the past few years, many saliency detection methods have been proposed based on the characteristics of the Human Visual System (HVS). Saliency detection methods in general can be categorized as either human eye fixation prediction~\cite{22Fang2012} approaches and salient object detection approaches~\cite{20Li2016}. The first one aims to identify salient locations where human observers fixate during scene view, and we call it as the eye fixation regions. The latter, salient object detection, focuses on predicting saliency values of pixels that determine whether the pixels belong to the salient object or not. In this paper, we focus on salient object detection task in video sequences.

Recently, deep learning \cite{20Li2016}, \cite{Zhang2017}, \cite{Huang2017} has been successfully utilized in object detection, semantic segmentation, object tracking and saliency detection. Despite recent great progress in saliency detection for still images, spatiotemporal saliency detection for video sequences remains challenging and it is much desired to design effective video saliency detection models. It is not easy to extract the accurate motion information in video sequences, and thus the small and fast moving objects in video sequences are usually difficult to be captured. Furthermore, the semantic properties of a visual scene are typically related to salient objects and the context close to these objects in this scene. Thus, how to extract and combine the temporal information and high-level spatial features such as semantic cues is important to design effective video saliency detection models.

Currently, there are many video saliency detection models proposed for various multimedia processing applications~\cite{52Kim2015}, \cite{53Liu2016}, \cite{54Wang2015}, \cite{56Fang2014}, \cite{Huang2017}-\cite{01Fang2014}. For traditional video saliency detection models, they first extract spatial and temporal features to compute spatial and temporal saliency maps, respectively; then the final saliency map for video sequences is predicted by combining the spatial and temporal saliency maps with certain fusion method \cite{53Liu2016}-\cite{56Fang2014}. Most of these methods manually extract low level features such as color, luminance and texture for spatial saliency estimation. However, they might loss some important high-level features such as semantic information in video sequences. What's more, some existing methods attempt to use linear or nonlinear combination rules to fuse spatial and temporal information simply \cite{52Kim2015}, \cite{53Liu2016}, \cite{54Wang2015}, \cite{56Fang2014}, which may ignore the intrinsic relationship due to the fixed weights used for the combination of spatial and temporal information.

In order to overcome these challenges, we adopt 3D convolutional and 3D deconvolutional neural networks to extract and fuse spatial and temporal features to build a effective video saliency detection model. In sum, the main contributions of the proposed method are summarized as follows:

1) We propose a novel saliency detection model for video sequences based on 3D convolutional neural networks. We construct a 3D convolutional network (Conv3DNet), which can be used to extract spatiotemporal features efficiently for saliency map prediction of video frames.

2) We devise a 3D deconvolutional network (Deconv3DNet) to learn saliency by fusing spatiotemporal features for the final saliency map calculation. Experimental results show that the proposed model outperforms other baseline methods on two large-scale datasets.

The rest of this paper is organized as follows: Section 2 give a detail description of the proposed deep saliency framework. Section 3 shows the comparison experimental results by using the state-of-the-art methods. Finally, we conclude this work in Section 4.

\section{Proposed Method}

\begin{figure*}
    \centering
    \includegraphics[width = 0.96\textwidth]{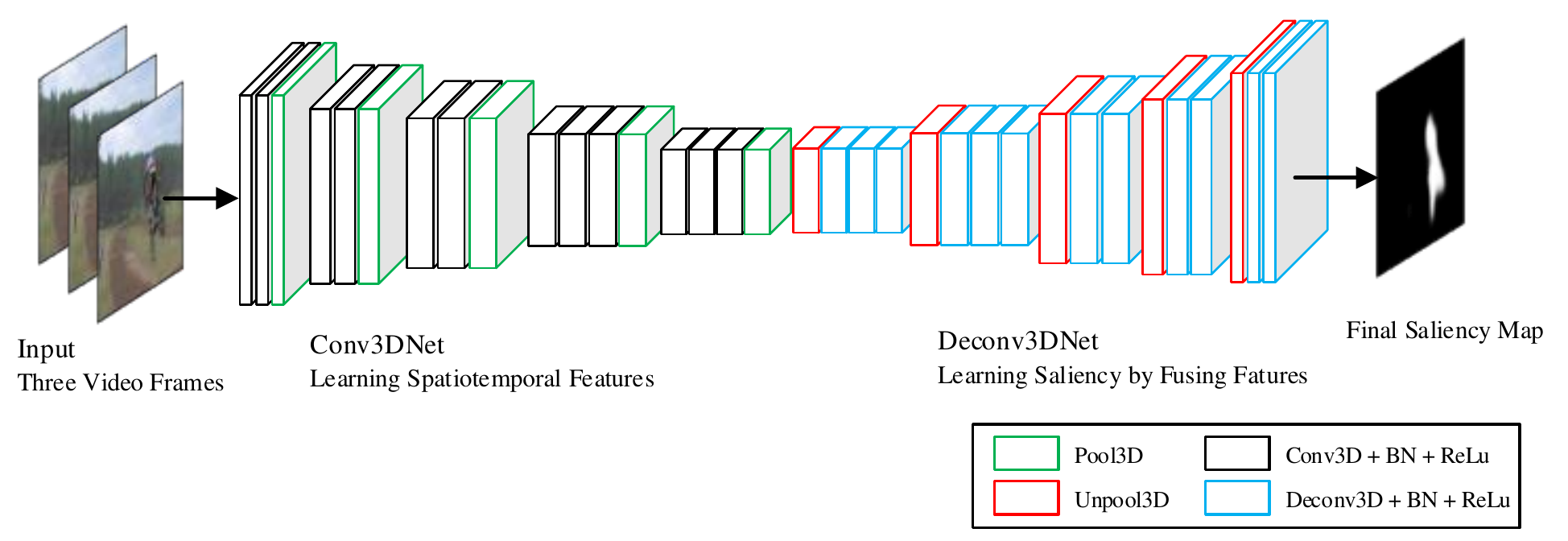}
    \caption{Architecture of the proposed video saliency detection model. There are two components in the proposed model: Conv3DNet with three consecutive video frames for spatiotemporal feature learning, and Deconv3DNet for the final saliency learning.}
\label{fig:framework}
\end{figure*}

The proposed model is demonstrated in Fig.~\ref{fig:framework}. As we can see from this framework, the proposed method includes two components: the Conv3DNet for spatiotemporal feature learning and the Deconv3DNet for saliency learning by fusing spatiotemporal features. With three consecutive video frames ($I_{t-1},I_{t},I_{t+1}$) in Conv3DNet, the ground truth map $G_{t}$ of video frame ($I_{t}$) in the training set is used to calculate the loss of forward propagation.

For simplicity, we denote $d \times k \times k$ as the kernel/stride size for 3D convolutional layer, 3D pooling layer, 3D deconvolutional layer and 3D unpooling layer, where $d$ represents the kernel/stride depth in temporal dimension and $k$ stands for the spatial filter/stride size. Besides, we intend to employ $f \times h \times w \times c$ to indicate the output shape of 3D convolution and deconvolution layers, where $f$ represents the number of input video frames; $h$, $w$, and $c$ are the parameters for height, width and channels of video frames or feature maps.

\subsection{The Spatiotemporal Stream Conv3DNet}

\begin{figure}
    \centering
    \includegraphics[width = 0.48\textwidth]{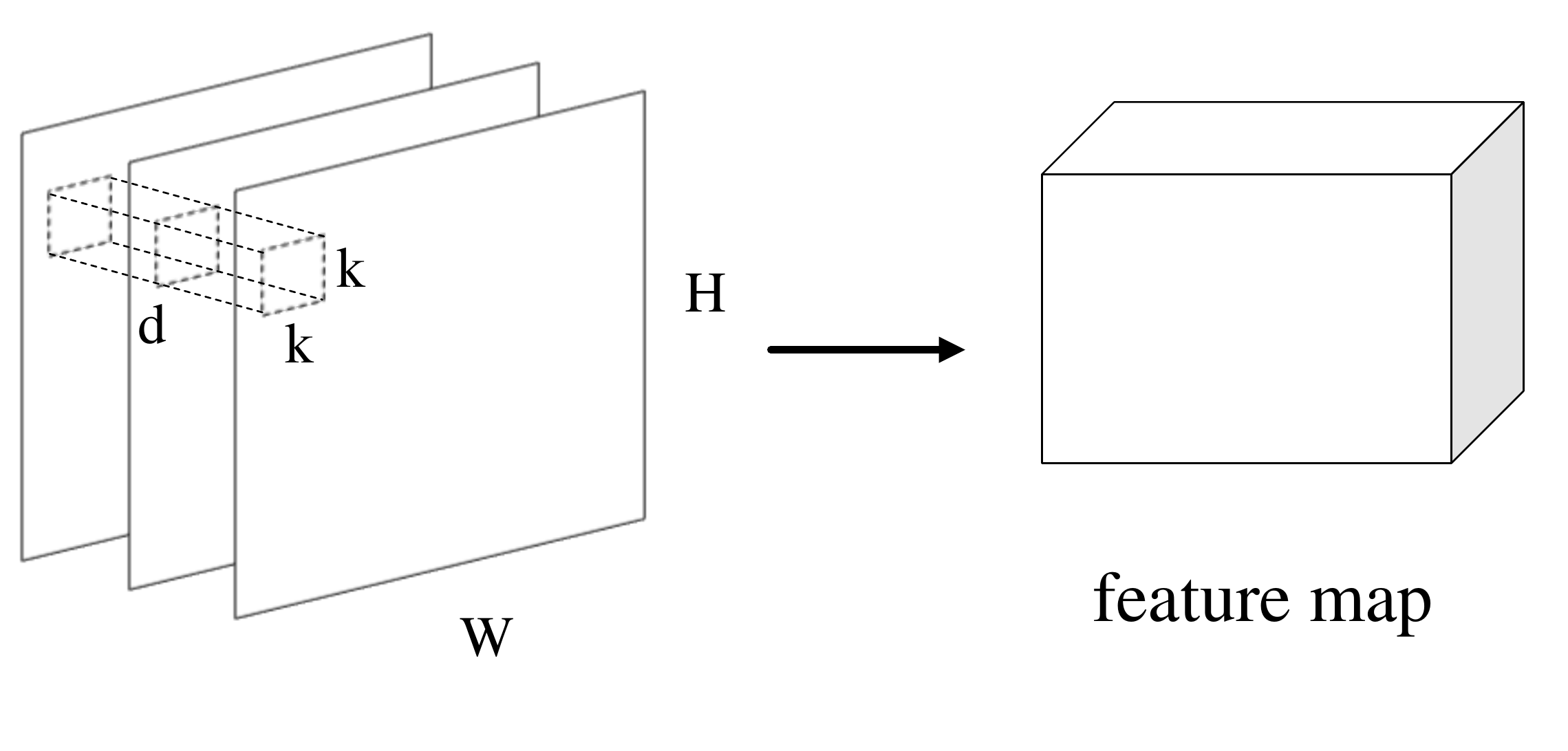}
    \caption{Illustration of 3D convolutional operation. The kernel of 3D convolutional layer is cube with size $d \times k \times k$, where $d$ represents the depth of temporal dimension and $k$ stands for the spatial filter size. $W$ and $H$ denote width and height of feature maps, respectively.}
\label{fig:3DConv}
\end{figure}

As shown in Fig.~\ref{fig:framework}, we construct a Conv3DNet including 5 group blocks, each of which consists of 3D pooling layer and 3D convolution layers with batch normalization and Relu (Rectified Linear Unites) operations. We design 5 group blocks for the proposed model, which would generate the feature maps with the size $7 \times 7$. The 3D convolutional operation is demonstrated in Fig.~\ref{fig:3DConv}. It can be used to learn spatiotemporal features simultaneously for video sequences. Moreover, Du \emph{et al.} demonstrated that 3D convolutional deep networks are useful and effective for learning spatiotemporal features~\cite{41Du2015}.

Due to the stride of convolutional and pooling operations, the output feature maps will be down-sampled and become sparse. This is the reason that we use Deconv3DNet to learn high-level temporal and spatial features. The Conv3DNet takes three consecutive video frames ($I_{t-1},I_{t},I_{t+1}$) as the input for learning the coherence and motion information between video frames, which have significantly contribute to video saliency detection. The output feature map $Y$ of convolutional operation can be denoted as follows:

\begin{eqnarray}
Y= f( \sum w \ast X + b )
\end{eqnarray}
where $\ast$ is convolutional operation, $X$ denote as the input feature map and $w$ represents the convolutional filter. After add a bias term $b$ to the convolutional results, they will be input an active function $f$ to improve the hierarchical nonlinear mapping learning capability.

Existing studies have shown that the convolutional filter with homogenous parameters of $3 \times 3 \times 3$ is effective for either 2D convolutional networks \cite{42Simonyan2015}, \cite{20Li2016} or 3D convolutional networks \cite{41Du2015}, thus we set 3D convolutional kernel as $3 \times 3 \times 3$ with strides $1 \times 1 \times 1$ in the proposed model. With direct extension of 2D max-pooling to the temporal field, many researches \cite{41Du2015} have demonstrated that 3D max-pooling operation can work on multiple temporal samples. As can be seen from Fig.~\ref{fig:framework}, the stride sizes of five 3D max-pooling layers are assigned as follows in the proposed method: $1 \times 2 \times 2$ for the first three max-pooling layers, $2 \times 2 \times 2$ for the last two max-pooling layers. We set these parameters for all 3D max-pooling layers as the above since we intend to learn more temporal features between video frames and do not expect to combine these temporal information at early stage.

\subsection{The Deconv3DNet for Saliency Learning}


As can be seen from Fig.~\ref{fig:framework}, we adopt 3D deconvolutional and unpooling operations in the proposed Deconv3DNet to fuse spatiotemporal features for video sequences and up-sample the resolution of feature maps. Specifically, we design a Deconv3DNet containing 5 group operations, including a set of 3D deconvolutional layers, 3D unpooling layers and batch normalization layers. As shown in Fig.~\ref{fig:framework}, we first provide a 3D unpooling layer, three 3D deconvolutional layers with batch normalization and Relu operation for the first and second group. Besides, we make use of one 3D pooling layer, two 3D deconvolutional layers for the last three groups, and all strides of the 3D unpooling layer are set as $1 \times 2 \times 2$ to upsample the spatial size of feature maps, while the temporal dimension is fixed to $1$ since we aim to calculate the saliency map of unitary frame $I_{t}$.

\begin{table}
\centering \caption{The basic information of four public datasets: DAVIS \cite{64Perazzi2016}, SegTrackV2 \cite{63Li2013}, VOT2016 \cite{Tomas2017}, USVD \cite{53Liu2016}.}
\begin{tabular}{  c | c | c | c }
\hline Datasets&Clips&Frames&Annotations\\
\hline\hline DAVIS \cite{64Perazzi2016}&50&3455&3455\\
\hline SegTrackV2 \cite{63Li2013}&14&1066&1066\\
\hline VOT2016 \cite{Tomas2017}&60&21646&21646\\
\hline USVD \cite{53Liu2016}&18&3550&3550\\

\hline\end{tabular}
\label{tab:Dataset}
\end{table}

At last, we utilize one extra 3D convolutional layer to generate the final saliency map with the size $224 \times 224 \times 1$ to keep high level saliency cues as much as possible. Here, we use Relu (Rectified Linear Unites) as the activation function of convolutional and deconvolutional layers. The square Euclidean error is used as the loss function. We denote ($I, G$) as a pair of training sampling, which consists of three frames $(I_{t-1}, I_{t}, I_{t+1})$ with the shape $h \times w \times 3$ and the corresponding ground truth map $G_{t}$ of the video frame $I_{t}$. Besides, we denote $S_{t}$ as the generated final saliency map. Because we intend to calculate saliency map of the single frame $I_{t}$, the goal of the proposed deep model is to optimize the following loss function on the mini-batch with size $k$:

\begin{eqnarray}
\emph{L}(S_{t},G_{t})= \frac{1}{k} \frac{1}{h} \frac{1}{w}	\sum_{l=1}^k \sum_{i=1}^h \sum_{j=1}^w \parallel S_{t}(i, j)-G_{t}(i, j)\parallel_{F}^2
\end{eqnarray}
where $S_{t}(i, j)$ and $G_{t}(i, j)$ denotes the pixel value of saliency map $S_{t}$ and ground truth map $G_{t}$.

Here, we adopt Adaptive Moment Estimation (Adam) \cite{Kingma2014} to optimize the proposed model. Adam is an optimization method that it uses the first moment estimation and second moment estimation of gradient to update the learning rate adaptively. During the training stage, all the parameters are learned by optimizing the loss function. More specifically, the loss function optimization aims to minimize the error between the saliency map generated by forward propagation and the corresponding ground truth map. In the test stage, the proposed model can predict the spatiotemporal saliency maps for any given video sequences without any prior knowledge by the trained model.

%
%
%

\section{Experimental Results and Analysis}

\subsection{Database and Evaluation Criteria}

In this study, we conduct the comparison experiments by using four public available benchmark video datasets: DAVIS \cite{64Perazzi2016}, SegTrackV2 \cite{63Li2013}, VOT2016 \cite{Tomas2017}, USVD \cite{53Liu2016}. We conclude the detailed information of these datasets in Table \ref{tab:Dataset}.
DAVIS \cite{64Perazzi2016} contains 50 natural video clips in total with diverse visual content including sports, car drift-turn, animals and outdoor video sequences, with various typical challenges such as multi moving objects, low contrast and complex background. SegTrackV2 \cite{63Li2013} consists of 14 video sequences with a variety of visual scenes and activities. VOT2016 \cite{Tomas2017} contains 60 video clips and 21646 corresponding ground truth maps with pixel-wise annotation of salient objects. USVD \cite{53Liu2016} contains 18 video sequences with binary ground truth maps that segment salient objects accurately for each video frame.

In our experiments, we utilize two datasets of VOT2016 \cite{Tomas2017} and USVD \cite{53Liu2016} to train the proposed video saliency detection model. To test the proposed video saliency model, we adopt the other two datasets DAVIS \cite{64Perazzi2016} and SegTrackV2 \cite{63Li2013} as the test data to evaluate the performance of the proposed method.

Similar with \cite{56Fang2014}, we report the quantitative performance evaluation results based on three popular metrics: Pearson’s Linear Correlation Coefficient (PLCC), Receiver Operating Characteristics (ROC) and Normalized Scanpath Saliency (NSS). PLCC is used to quantify the correlation and dependence, demonstrating statistical relationship between the saliency maps and ground truth maps. PLCC is commonly defined as follows:

\begin{eqnarray}
PLCC(s,f)=\frac{cov(s,f)}{\sigma_s\sigma_f}
\end{eqnarray}
where $cov(s,f)$ denotes the covariance of saliency map $s$ and ground truth map $f$; $\sigma_s$ and $\sigma_f$ stand for the standard deviation values of the saliency map $s$ and ground truth map $f$, respectively. The range of PLCC values is [0,1]. Obviously, the lager PLCC value means the better performance of the saliency detection model.

In addition, ROC curve and area under ROC curve (AUC) are also used for evaluating the performance of saliency detection models. With the varied threshold, the ROC curve is plotted as the False Positive Rate (FPR) and True Positive Rate (TPR), which are defined as follows:

\begin{eqnarray}
  FPR = \frac{M \cap \bar{G}}{\bar{G}}      \\
  TPR = \frac{M \cap G}{G}
\end{eqnarray}
where $M$ represents the binary mask of the saliency map generated by a series of varying discrimination thresholds on the saliency map; $G$ denotes the binary ground truth map while $\bar{G}$ stands for the reverse of $G$. Generally, the lager AUC value means the better performance of saliency detection model.

As a widely adopted to evaluate the saliency detection method, NSS is defined by the response value at human fixation locations in the normalized saliency map with zero mean and unit standard deviation as:

\begin{eqnarray}
NSS(s,g)=\frac{1}{\sigma_s}(s(g_i,g_j)-\mu_s)
\end{eqnarray}
where $s$ and $g$ denote the saliency map and corresponding ground truth map; $(g_i,g_j)$ is the pixel location of the ground truth map; $\mu_s$ and $\sigma_s$ represent the mean value and the standard deviation of the saliency map, respectively. Typically, the higher NSS value means better saliency detection model.

In our experiments, the proposed deep network of video salient object detection is implemented in Ubuntu operating system with the toolbox, Tensorflow library \cite{65Abadi2016}, an open source software for deep learning developed by Google. We use a computer with Intel Core I7-6900K*16 CPU (3.20GHz), 64 GB RAM and Nvidia TITAN X (Pascal) GPU with 16 GB memory.

\subsection{Performance Comparison}

Furthermore, we compare the proposed approach against several existing video saliency detection methods including Fang \cite{56Fang2014}, LGGR \cite{54Wang2015}, MultiTask \cite{20Li2016}, RWRV \cite{52Kim2015}, CE \cite{0Li2009} and SGSP \cite{53Liu2016}. We show the quantitative experimental results in Table \ref{tab:compare on Seg} on SegTrackV2 dataset \cite{63Li2013} and Table \ref{tab:compare on DAVIS} on DAVIS dataset \cite{64Perazzi2016}, where the PLCC, AUC and NSS scores are collected from the mean value of 14 video sequences in SegTrackV2 dataset and 50 video sequences in DAVIS dataset, respectively.

Among these state-of-the-art approaches, Fang \cite{56Fang2014} is a Gestalt theory based saliency detection method; LGGR \cite{54Wang2015} uses local gradient flow optimization and global refinement for video saliency prediction; MultiTask \cite{20Li2016} is a deep learning based salient object detection method for images, using multi-tasks of saliency detection and semantic segmentation; RWRV \cite{52Kim2015} predicts video saliency via random walk with restart method; CE \cite{0Li2009} is a video saliency computation approach based on conditional entropy; SGSP \cite{53Liu2016} utilizes superpixel-level graph and spatiotemporal propagation method for saliency detection.

\begin{figure*}
    \centering
    \includegraphics[width=0.48\textwidth]{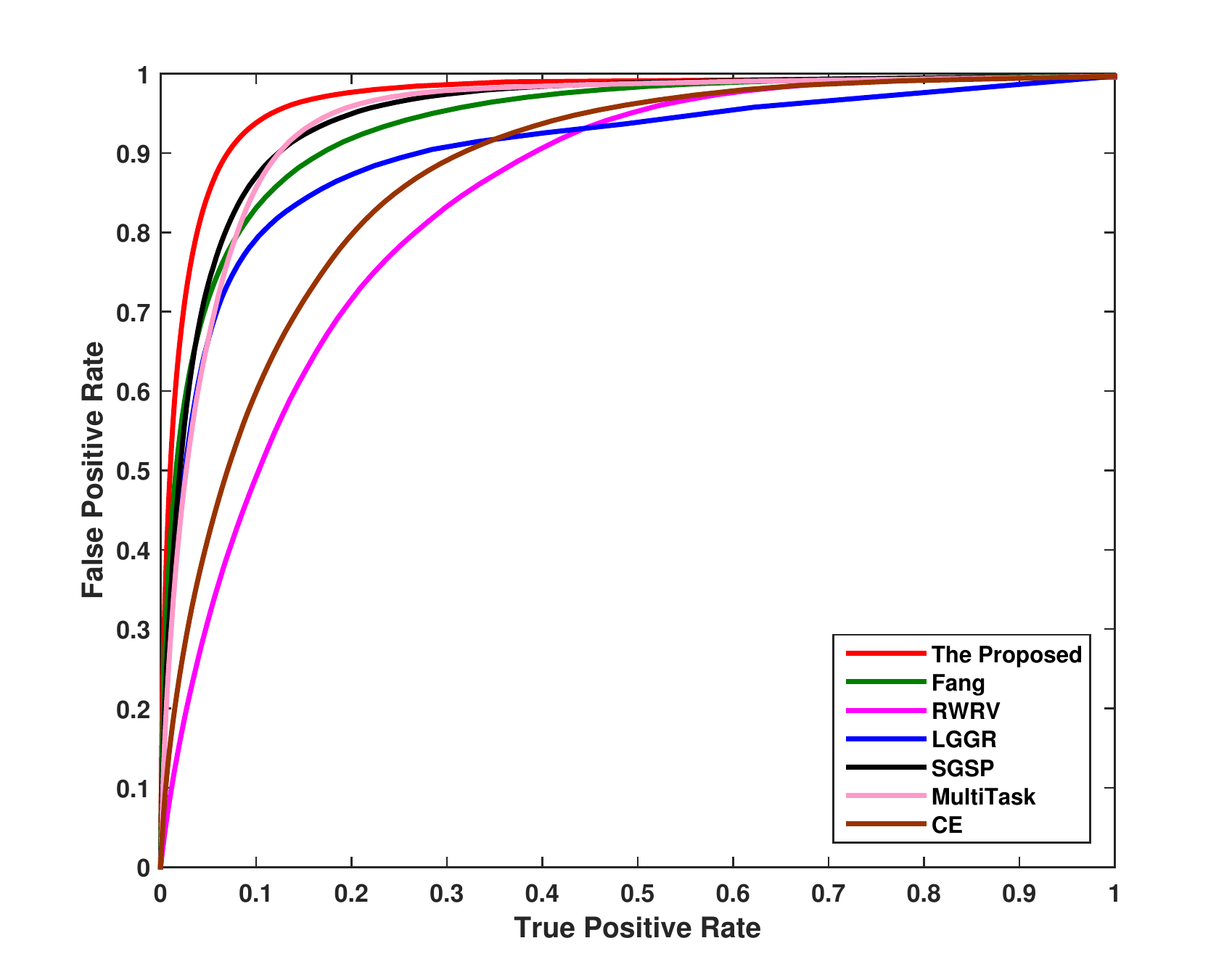}
    \includegraphics[width=0.48\textwidth]{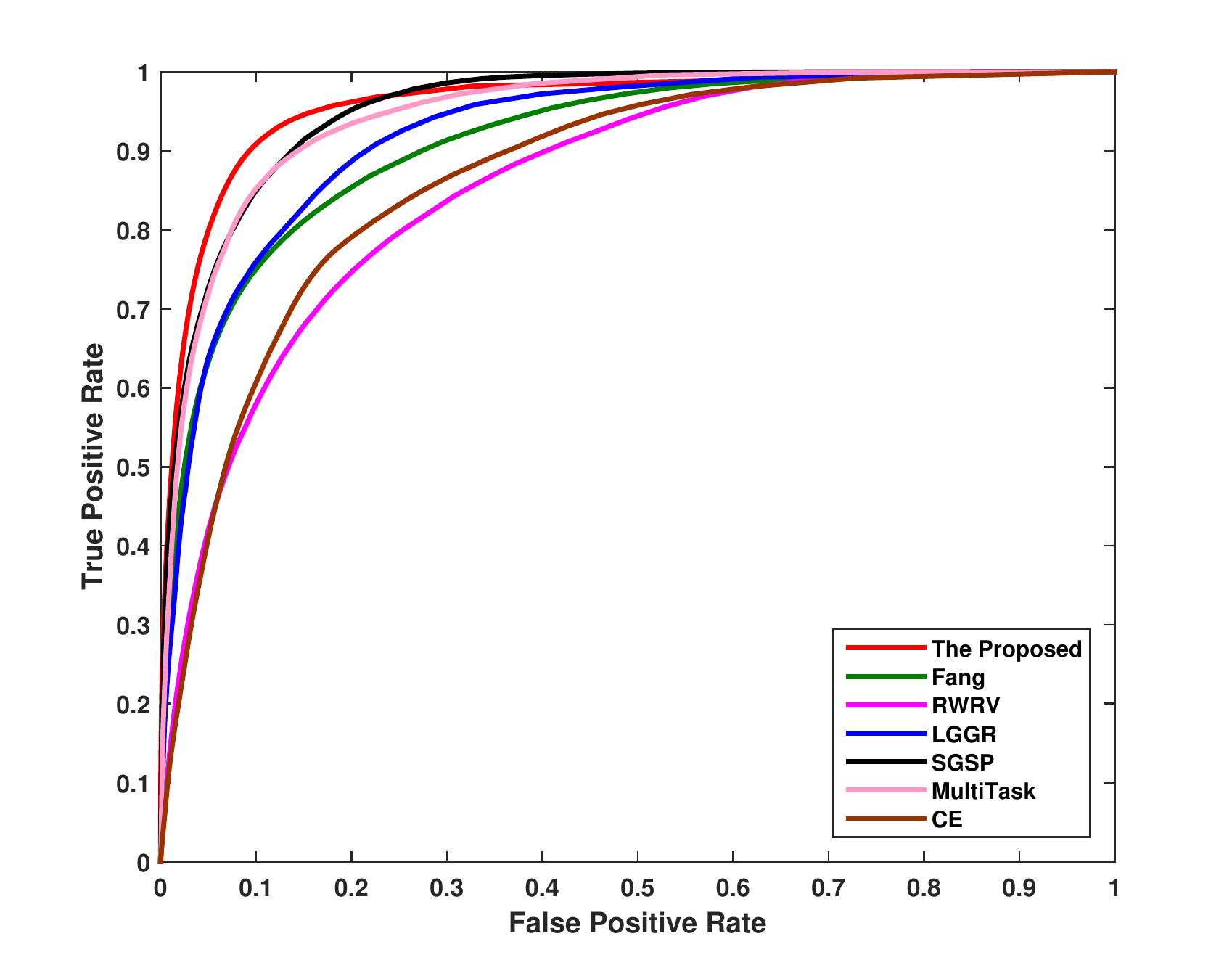}
    \caption{ROC comparison of spatiotemporal saliency models on SegTrackV2 dataset \cite{63Li2013} (left) and DAVIS dataset \cite{64Perazzi2016} (right).}
\label{fig:ROC SegTractV2}
\end{figure*}

\begin{figure*}
    \centering
    \includegraphics[width=0.96\textwidth]{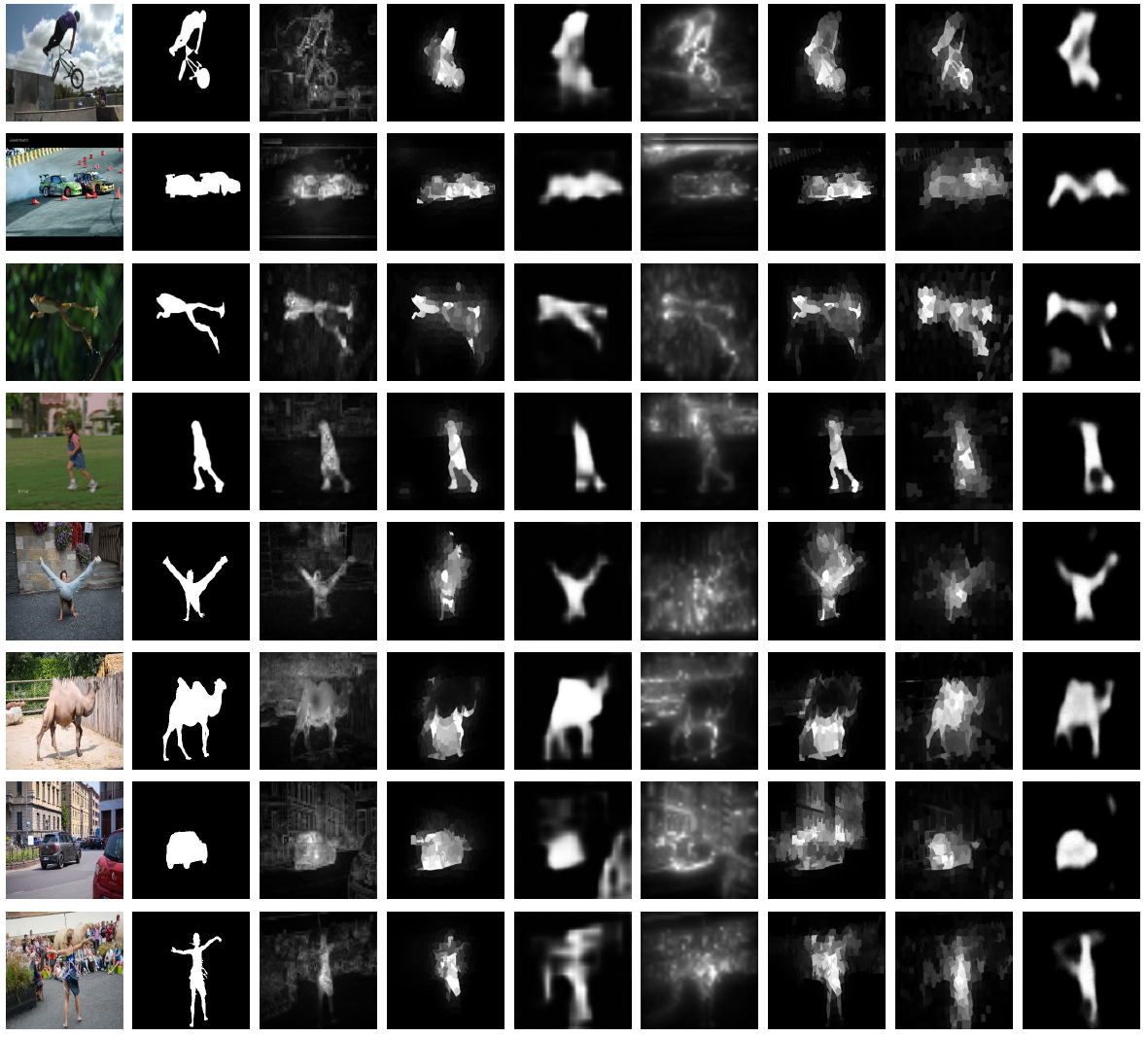}
    \caption{Comparison of different video saliency detection models on dataset SegTrackV2 \cite{63Li2013} (first four rows) and DAVIS \cite{64Perazzi2016} (last four rows). First column to the final column: original video frames; the ground truth maps; saliency maps from Fang \cite{56Fang2014}, LGGR \cite{54Wang2015}, MultiTask \cite{20Li2016}, RWRV \cite{52Kim2015}, CE \cite{0Li2009}, SGSP \cite{53Liu2016} and the proposed method.}
\label{fig:compare}
\end{figure*}

We provide some visual saliency samples from different saliency detection models in Fig. \ref{fig:compare} on SegTrackV2 \cite{63Li2013} (first four rows) and DAVIS \cite{64Perazzi2016} (last four rows). It can be seen that the saliency map obtained from other existing methods contain some noises, as shown by the fact that some background regions are detected as the salient locations in some saliency maps generated from existing methods. Although the saliency maps of LGGR \cite{54Wang2015} and SGSP \cite{53Liu2016} have relatively better results than the RWRV \cite{52Kim2015} method, however, those models still existing fatal block effect and loses some visually important information in the saliency map since they divided the video frames into block/super-pixel to calculate the local/globle feature contrast.

\begin{table}
\centering \caption{Comparison of different video saliency detection models on SegTrackV2 \cite{63Li2013}.}
\begin{tabular}{  c | c | c | c | c | c | c | c }
\hline
Models& Fang & LGGR & MultiTask & RWRV & CE & SGSP & Proposed Model \\
\hline\hline PLCC & 0.5098 & 0.7133 & 0.7752 & 0.5831 & 0.4595 & 0.6452 & \textbf{0.7838}\\
\hline AUC  & 0.7936 & 0.8887 & 0.9099 & 0.8504 & 0.8257 & 0.8660 & \textbf{0.9107}\\
\hline NSS & 2.5876 & 2.5895 & 3.0762 & 2.0302 & 1.8046 & 2.9739 & \textbf{3.0830}\\
\hline\end{tabular}
\label{tab:compare on Seg}
\end{table}

\begin{table}
\centering \caption{Comparison of different video saliency detection models on DAVIS \cite{64Perazzi2016}.}
\begin{tabular}{  c | c | c | c | c | c | c | c  }
\hline
Models& Fang & LGGR & MultiTask & RWRV & CE & SGSP & Proposed Model \\
\hline\hline PLCC & 0.6720 & 0.6733 & 0.8138 & 0.4077 & 0.4985 & 0.7439 & \textbf{0.8145}\\
\hline AUC  & 0.9034 & 0.8735 & 0.9262 & 0.8282 & 0.8436 & 0.9114 & \textbf{0.9325}\\
\hline NSS & 2.5904 & 2.4775 & 2.8294 & 1.6699 & 1.7874 & 2.7747 & \textbf{2.9485}\\
\hline\end{tabular}
\label{tab:compare on DAVIS}
\end{table}

Meanwhile,  From both Tables \ref{tab:compare on Seg} and \ref{tab:compare on DAVIS}, we can observe that the proposed method can obtain better video saliency prediction performance than other related ones, as shown by the highest PLCC, AUC and NSS values among the compared models. We provide the ROC curves of all these methods in Fig.~\ref{fig:ROC SegTractV2} on SegTrackV2 dataset \cite{63Li2013} (left) and DAVIS dataset \cite{64Perazzi2016} (right) to demonstrate the better results of our model than other existing ones.

\section{Conclusion}

In this paper, a novel salient object detection approach with 3D convolutional neural networks is proposed to effectively learn semantic and spatiotemporal features for video sequences. The proposed model mainly includes two components: the spatiotemporal stream Conv3DNet and the Deconv3DNet for saliency learning. The Conv3DNet consists of a series of 3D convolutional layers, which is proved to be effective to obtain spatiotemporal information between consecutive frames ($I_{t-1},I_{t},I_{t+1}$). The Deconv3DNet is designed to combine the spatiotemporal features from Conv3DNet to learn the final spatiotemporal saliency map for video. Experimental results have shown that there is great potential to build the video saliency detection model with 3D convolutional operation for effectively learning spatiotemporal features instead of time-consuming hand-crafted features.

\subsection*{Acknowledgments.}
This work was supported by NSFC (No. 61571212), and NSF of Jiangxi Province in China (No. 20071BBE50068, 20171BCB23048, 20161ACB21014, GJJ160420).


\end{document}